\PassOptionsToPackage{table}{xcolor}
\documentclass{article}
\usepackage[table]{xcolor}

\usepackage{PRIMEarxiv}

\usepackage[utf8]{inputenc} 
\usepackage[T1]{fontenc}    
\usepackage{hyperref}       
\usepackage{url}            
\usepackage{booktabs}       
\usepackage{amsfonts}       
\usepackage{nicefrac}       
\usepackage{microtype}      
\usepackage{lipsum}
\usepackage{fancyhdr}       
\usepackage{graphicx}       
\graphicspath{{media/}}     

\usepackage{subcaption}
\usepackage{float}
\usepackage{tcolorbox}
\usepackage{balance} 
\usepackage{amsmath} 

\newcommand{\datasetname}{S-HArM}

\title{``Humor, Art, or Misinformation?'': A Multimodal Dataset for Intent-Aware Synthetic Image Detection
\thanks{
Accepted at the Diffusion of Harmful Content on Online Web (DHOW) Workshop at the 33rd ACM International Conference on Multimedia (ACM MM), 2025. DOI: \url{https://doi.org/10.1145/3746275.3762215}}}

\usepackage{authblk}

\author[1]{\small Anastasios Skoularikis}
\author[1, 2]{\small Stefanos-Iordanis Papadopoulos\thanks{Corresponding author} \ }
\author[2]{\small Symeon Papadopoulos}
\author[1]{\small Panagiotis C. Petrantonakis}

\affil[1]{\footnotesize Department of Electrical \& Computer Engineering, Aristotle University of Thessaloniki.}
\affil[2]{\footnotesize Information Technology Institute, Centre for Research \& Technology Hellas.}
\affil[ ]{
\textit{skouanas@ece.auth.gr},
\textit {\{stefpapad,papadop\}@iti.gr, \textit{ppetrant@ece.auth.gr}}}

\begin{document}
\maketitle

\begin{abstract}
Recent advances in multimodal AI have enabled progress in detecting synthetic and out-of-context content.
However, existing efforts largely overlook the intent behind AI-generated images.
To fill this gap, we introduce \textit{S-HArM}, a multimodal dataset for intent-aware classification, comprising 9,576 ``in the wild'' image–text pairs from Twitter/X and Reddit, labeled as \textit{Humor/Satire}, \textit{Art}, or \textit{Misinformation}.
Additionally, we explore three prompting strategies (image-guided, description-guided, and multimodally-guided) to construct a large-scale synthetic training dataset with Stable Diffusion. 
We conduct an extensive comparative study including modality fusion, contrastive learning, reconstruction networks, attention mechanisms, and large vision-language models. 
Our results show that models trained on image- and multimodally-guided data generalize better to ``in the wild'' content, due to preserved visual context. 
However, overall performance remains limited, highlighting the complexity of inferring intent and the need for specialized architectures. 
\end{abstract}


\section{Introduction}

\begin{figure*}[t]
    \centering
    \includegraphics[width=0.9\linewidth]{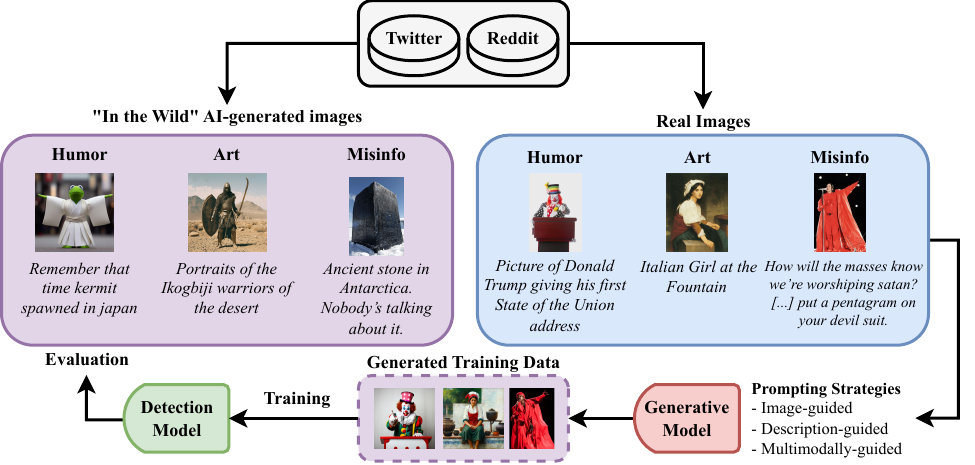}
    \caption{Overview of the proposed pipeline. 
    We collect ``in the wild'' AI-generated images 
    to build the \textit{\datasetname} evaluation benchmark, and real images to generate synthetic training data via Stable Diffusion using three prompting strategies. Various detection models are trained on synthetic training data and evaluated on the evaluation benchmark.
    }
    \label{fig:teaser}
\end{figure*}

The rise of generative models has significantly affected the landscape of digital media, enabling the large-scale and rapid production of realistic synthetic images that are becoming increasingly indistinguishable from real ones. 
From the inception of Generative Adversarial Networks (GANs) and their variants \cite{jabbar2021survey} to recent advances in diffusion models \cite{yang2023diffusion}, synthetic image generation has become more powerful and accessible.
The wide availability of tools such as Adobe Firefly, DALL$\cdot$E, and Midjourney allow users to create realistic content with minimal effort and little to no technical expertise. 
While these tools can foster creative expression, particularly in art and satire, they also raise concerns over the misuse of synthetic content, with potentially serious social, political, and economic consequences \cite{karnouskos2020artificial}.

Researchers have responded by developing numerous methods for synthetic image detection \cite{wang2020cnngeneratedimagessurprisinglyeasy, sinitsa2024deepimagefingerprintlow, 
synthbuster, 
Li_2024_CVPR, konstantinidou2025texturecropenhancingsyntheticimage,
wu2025generalizablesyntheticimagedetection, 
rine, 
karageorgiou2025any} 
and identifying multimodal forms of misinformation \cite{
verif2015, 
Boididou2018TwitterVerification,
jin2016novel,
nakamura2020rfakeddit,
aneja2023cosmos, 
luo2021newsclippingsautomaticgenerationoutofcontext, 
sabir2018deep, 
mu2023self,
papadopoulos2023synthetic,
abdelnabi2022open,
yuan2023support,
papadopoulos2024reddotmultimodalfactcheckingrelevant,
zhang2023ecenet,  
qi2024sniffermultimodallargelanguage,
Papadopoulos_2025_WACV}. 
These efforts largely focus on detecting AI-generated,  manipulated images, or decontextualized images; and have made significant progress in recent years. 
Nevertheless, these approaches often overlook a crucial aspect: the intent behind the creation of synthetic images; whether the image was meant as artistic expression, humor, or deliberate misinformation.

Intent classification can serve as a valuable first step in moderation pipelines by prioritizing harmful content (e.g., misinformation) for human review over benign cases like satire or art.
Crucially, intent cannot always be inferred from visual content alone. 
An image generated for satire or creative purposes may be visually indistinguishable from one designed to mislead. 
To capture these subtle distinctions, a multimodal approach, integrating both visual and textual signals, is essential for uncovering the creator's underlying intention.
For example, consider Fig. \ref{fig:teaser}, where an AI-generated image is accompanied by the user post: ``Ancient stone in Antarctica. Nobody's talking about it. I wonder why.'' 
The claim is factually incorrect and implies a baseless conspiracy and the AI-generated image is used as ``evidence'' to support it. 
However, without the textual modality, the intent behind the generation of this image would be ambiguous. 

To address this gap, we introduce \textit{\datasetname} (Synthetic–Humor, Art, Misinformation), a multimodal dataset for intent-aware classification of AI-generated images. 
Unlike existing datasets that focus on binary classification (i.e., real vs. generated), \textit{\datasetname} supports a three-way classification task: humor/satire, art, or misinformation.
\textit{\datasetname} comprises two components:
(1) an annotated ``in the wild'' evaluation benchmark of AI-generated image created and shared by users in social media platforms, 
and (2) a larger training set of synthetic data generated from real images using Stable Diffusion.
To build the evaluation benchmark, we curated content from Twitter/X's Community Notes—identifying examples of misinformation and satire—and from selected subreddits focused on art and humor.
For training data, we collected additional Twitter/X and Reddit posts with real (non-generated) images, which were used to generate synthetic versions using three generative prompting strategies:
\textit{Image-guided}: the real image serves as the primary input to guide generation, 
\textit{Description-guided}: a captioning model first produces a textual description of the real image, which is then used as input for text-to-image generation;
\textit{Multimodally-guided}: both the real image and its generated description are combined to jointly guide generation.
In total, \textit{\datasetname} includes 9,576 annotated evaluation image-text pairs, and over 87,000 synthetic training samples per generative prompting strategy. 

Furthermore, we propose an end-to-end pipeline and conduct a comprehensive comparative study for intent-aware classification.
As illustrated in Fig.~\ref{fig:teaser}, after collecting data and constructing both the ``in the wild'' evaluation benchmark and three versions of the training set,
we train a broad range of models, including:
unimodal baselines using either image-only or text-only inputs, multimodal baselines,
contrastive learning,
reconstruction networks,
attention-based models,
and large vision–language models (LVLMs), used as zero-shot classifiers.
Finally, we evaluate the generalization of models trained on synthetic data to ``in the wild'' AI-generated content. 

Our results show that models trained on data generated through image-guided and multimodally-guided strategies tend to generalize better due to the preservation of crucial visual characteristics.
While performance on the synthetic validation set is very high (96.6\%), generalization to ``in the wild'' content remains limited, with the best model reaching 71.6\% accuracy.
This attests the difficulty of intent-aware classification, which requires a deeper understanding of abstract concepts such as humor, misinformation, and artistic expression—often conveyed through subtle cross-modal cues. 
Finally, our empirical evaluations confirm that the task needs to be addressed multimodally, since unimodal baselines (image-only or text-only) consistently underperformed, suggesting that intent emerges from the interaction between visual and textual modalities.
We share the code and dataset at: \url{https://github.com/Qedrigord/SHARM}.

\section{Related Work}

The rise of generative AI has enabled both creative expression and harmful misuse. While these tools empower users to create digital art, satire, and memes, they are increasingly exploited for misinformation and manipulation \cite{misinfo_ai, NGUYEN2022103525}. 
Within algorithmically driven echo chambers, personalized content reinforces existing beliefs, amplifying polarization and radicalization—especially due to cognitive biases like confirmation bias and the continued influence effect \cite{misinfo_ai, muhammed2022disaster}.
AI-generated images amplify this risk by enabling personalized, deceptive narratives \cite{aimeur2023fake}, fueling the rise of multimodal misinformation that blends synthetic visuals and text to mislead more effectively \cite{verif2015}. 
Deepfakes add further threats, including fraud, privacy violations, and political misuse. As manual moderation becomes unsustainable, scalable, automated, and intent-aware detection is urgently needed to protect digital trust \cite{NGUYEN2022103525}.

\subsection{Multimodal Misinformation Detection}

Given the significant social impact of misinformation and the growing prevalence of multimodal forms of misinformation,
extensive research has focused on developing effective detection methods, and particularly on Out-of-Context (OOC) misinformation. 
Recent approaches have leveraged fine-tuned CLIP embeddings \cite{luo2021newsclippingsautomaticgenerationoutofcontext}, self-supervised distillation frameworks \cite{mu2023self}, and Transformer-based encoders for modeling joint cross-modal attention \cite{papadopoulos2023synthetic}.
Other lines of work incorporate external evidence retrieved from the Web, using models that assess consistency \cite{abdelnabi2022open}, 
stance \cite{yuan2023support} or relevance of external evidence towards the image-text pair under examination  \cite{papadopoulos2024reddotmultimodalfactcheckingrelevant}.
Furthermore, there is also growing interest in using LVLMs due to their enhanced reasoning capabilities and potential for explainability \cite{zhang2023ecenet, qi2024sniffermultimodallargelanguage}.
Nevertheless, a recent study has raised critical concerns regarding current benchmarks for OOC detection, showing that models often exploit dataset-specific artifacts rather than genuinely reasoning about factuality \cite{Papadopoulos_2025_WACV}.

\subsection{Multimodal Misinformation Datasets}

To support research on multimodal misinformation detection, a range of datasets have been introduced.
Early efforts include the MediaEval 2015 and 2016 Twitter datasets \cite{verif2015, Boididou2018TwitterVerification}, as well as the Weibo dataset \cite{jin2016novel}, which contain social media posts annotated as real or fake based on whether the accompanying multimedia accurately reflects the described event.
Due to their limited size and manual annotation effort, later works adopted large-scale weak supervision. 
For example, Fakeddit \cite{nakamura2020rfakeddit} includes over a million Reddit posts weakly labeled according to the originating subreddit.
Other approaches generate synthetic misinformation through algorithmic means. 
COSMOS \cite{aneja2023cosmos} randomly samples mismatched image–text pairs to simulate OOC misinformation. 
NewsCLIPpings \cite{luo2021newsclippingsautomaticgenerationoutofcontext} builds on legitimate image–text pairs from VisualNews \cite{liu2021visualnewsbenchmarkchallenges}, perturbing them via intra- and cross-modal similarity techniques to produce more realistic OOC samples. 
In contrast, the Multimodal Entity Image Repurposing (MEIR) introduce misinformation by manipulating named entities in textual content \cite{sabir2018deep}.
To evaluate the generalization of models trained on weakly annotated or synthetically generated datasets—and to mitigate unimodal biases— VERITE \cite{Papadopoulos2024VERITE} was introduced as a robust benchmark for multimodal misinformation detection.

\subsection{Synthetic Image Detection}

Beyond multimodal misinformation detection, there has been growing interest in the detection of synthetic images. 
Early efforts relied on CNN-based detectors, evaluating their ability to generalize across various generative architectures \cite{wang2020cnngeneratedimagessurprisinglyeasy, sid}. 
Other works targets artifacts introduced by CNN-based generators such as GANs and Diffusion models, either in the spatial domain \cite{sinitsa2024deepimagefingerprintlow} or frequency domain \cite{synthbuster, Li_2024_CVPR}.  Similarly, TextureCrop \cite{konstantinidou2025texturecropenhancingsyntheticimage} enhances detection performance by extracting high-frequency image patches where generative artifacts are concentrated. 
Contrastive and self-supervised learning approaches are also widely used—for example, using InfoNCE loss to align image and text representations \cite{wu2025generalizablesyntheticimagedetection}, or leveraging intermediate CLIP features to capture fine-grained inconsistencies \cite{rine}, or self-supervised spectral reconstruction \cite{karageorgiou2025any}.

\subsection{Synthetic Image Datasets}

To support the development and evaluation of synthetic image detection methods, several datasets have been introduced. Both the ARTIFACT dataset \cite{artifact} and Corvi et al. \cite{Corvi} use the COCO dataset \cite{coco} as a base to generate synthetic images across diverse categories, employing both Generative Adversarial Networks (GANs) and Diffusion Models. In contrast, Forensynths \cite{wang2020cnngeneratedimagessurprisinglyeasy} relies exclusively on GANs and draws from a more varied set of source datasets. Datasets based solely on Diffusion Models include Synthbuster \cite{synthbuster}, which used the Raise \cite{Raise} and Dresden \cite{Dresden} datasets to construct prompts, and CIFAKE \cite{cifake}, which generated images using CIFAR-10–inspired prompts to mimic the original dataset. A different approach is taken by TWIGMA \cite{twigma}, which curated AI-generated images directly from Twitter using specific hashtags. 
While these datasets provide valuable resources for synthetic image detection, they lack annotations related to  humor/satire, art, or misinformation, making them unsuitable for our task.

\section{Construction of \datasetname}

The Synthetic Humor, Art, or Misinformation (\textit{\datasetname}) dataset consists of two main components: an annotated ``in the wild'' evaluation benchmark and a synthetic large-scale training set.

\subsection{``In the Wild'' Evaluation Benchmark}
To construct the evaluation benchmark, we collected real-world, or ``in the wild'', examples of AI-generated images created and shared by users on two major social media platforms: Twitter/X and Reddit. 

\subsubsection{Reddit data}
Reddit data was sourced from publicly available dumps, obtained using Arctic Shift\footnote{\url{https://github.com/ArthurHeitmann/arctic_shift}}. 
To ensure the synthetic origin of the images, we focused exclusively on subreddits dedicated to content generated by AI-based models. 
We applied several quality control measures to ensure thematic consistency and data relevance:
(1) Only subreddits with active moderation and clear thematic focus were considered, as off-topic posts are typically removed by community moderators.
(2) \textit{High approval threshold}: Posts were filtered for an upvote-ratio above 0.9, indicating strong community endorsement and alignment with subreddit themes.
(3) When available, ``post flairs'' (tags and labels by users and/or moderators) were employed to assist in identifying content categories.
Additional filtering criteria ensured consistency and linguistic relevance: posts had to be in English; each post had to contain an image; post titles were required to have at least four words.
For each candidate subreddit, 20 randomly selected posts passing the above filters were manually reviewed to verify content quality and relevance. 
Subreddits with frequent low-quality or off-topic content were excluded.
Data collection was carried out in January 2025. 
Table \ref{tab:reddit_data} shows 
the number of collected image-post pairs collected from Reddit regarding \textit{Humor/Satire} and \textit{Art}. 
We were not able to identify any subreddits to consistently represent the \textit{Misinformation} category.

\begin{table}
  \caption{Number of collected posts from subreddits associated with AI-generated images of \textit{Art} and \textit{Humor/Satire}. 
  Parentheses indicate ``flairs'': user- or moderator-assigned tags.
  }
  \label{tab:reddit_data}
  \centering
  \begin{tabular}{lc}
    \toprule

    \textbf{Humor/Satire Subreddit} & \textbf{Posts}\\
    \midrule
    aigeneratedmemes & 2,552 \\
    aimemes & 599 \\
    hellaflyai & 10,764 \\
    midjourney (Jokes/Meme – Midjourney AI) & 1,281 \\
    StableDiffusion (Meme) & 1,934 \\

    \midrule

        \textbf{Art Subreddit} & 
        \\
    \midrule
    midjourney (AI Showcase – Midjourney) & 8,202 \\
    StableDiffusion (Workflow Not Included) & 8,918 \\

    \bottomrule
  \end{tabular}
\end{table}

\subsubsection{Twitter data}
To collect data from Twitter/X, we leveraged the publicly available Community Notes\footnote{\url{https://communitynotes.x.com}}, a crowdsourced framework for assessing whether tweet content is misleading or satirical. 
Each note includes binary field indicators\footnote{Misinformation fields: `misleadingManipulatedMedia', `misleadingFactualError', `misleadingOutdatedInformation', `misleadingMissingImportantContext', `misleadingUnverifiedClaimAsFact', `misleadingOther', Humor/Satire fields: `misleadingSatire', `notMisleadingClearlySatire'} for misinformation and satire. 
Notes also include a free-text summary where contributors justify their assessments.

For the \textit{Misinformation} class, we selected tweets with at least one misinformation indicator (field) marked as true and both satire indicators were marked as false. 
For the \textit{Humor/Satire} class, we included tweets with at least one satire-related indicator marked as true. 
In both cases, we further filtered for tweets whose summary text contained keywords related to generative artificial intelligence (AI, Artificial Intelligence, A.I.); ensuring that the image was likely AI-generated. 
For the \textit{Art} category, we used the Art Community Generative AI \footnote{\url{https://x.com/i/communities/1601841656147345410}}, a Twitter/X community dedicated to AI-generated artistic content. 
To ensure relevance and engagement, we retained only posts with a minimum of five `favourites\_count', as a proxy for user approval.
All posts were required to be in English and include at least one image. 
Data collection took place in January 2025. 
The amount of data collected from Twitter/X is summarized in the second column of Table \ref{tab:testtotalbal}

\subsubsection{Evaluation Benchmark Summary}
Our data collection process yielded numerous samples for the \textit{Art} and \textit{Humor/Satire} categories from Reddit, 
and \textit{Art} from Twitter, 
whereas the \textit{Misinformation} class was significantly underrepresented and sourced exclusively from Twitter. 
To address this class imbalance, we applied random under-sampling to Reddit-derived \textit{Art} and \textit{Humor/Satire}, and Twitter-derived \textit{Art},  
resulting in 3,192 samples per class, as summarized in Table~\ref{tab:testtotalbal}.  
Due to its curated nature and limited size, this set is meant for use exclusively in evaluation.

\begin{table}
  \caption{Number of ``in the wild'' posts (image–text pairs) from Reddit and Twitter after balancing.}
  \label{tab:testtotalbal}
  \centering
  \begin{tabular}{lccc}
    \toprule
    \textbf{Category} & \textbf{Twitter} & \textbf{Reddit} & \textbf{Total} \\
    \midrule
    Humor/Satire & 436 & 2,756 & 3,192 \\
    Art & 1,596 & 1,596 & 3,192 \\    
    Misinformation & 3,192 & – & 3,192 \\
    \bottomrule
  \end{tabular}
\end{table}

\begin{table}
  \caption{Number of posts with real (non-synthetic) images from Reddit and Twitter after balancing.}
  \label{tab:trainbal}
  \centering
  \begin{tabular}{lccc}
    \toprule
    \textbf{Category} & \textbf{Twitter} & \textbf{Reddit} & 
    \textbf{Total} \\
    \midrule
    
    Humor/Satire & 3,789 
    & 25,385 
    & 29,174 \\    
    
    Art & – & 
    29,174 &  
    29,174 \\
        
    Misinformation 
    & 29,174 
    & – 
    & 29,174 \\
    \bottomrule
  \end{tabular}
\end{table}

\begin{figure*}[t]
    \centering
    \includegraphics[width=\linewidth]{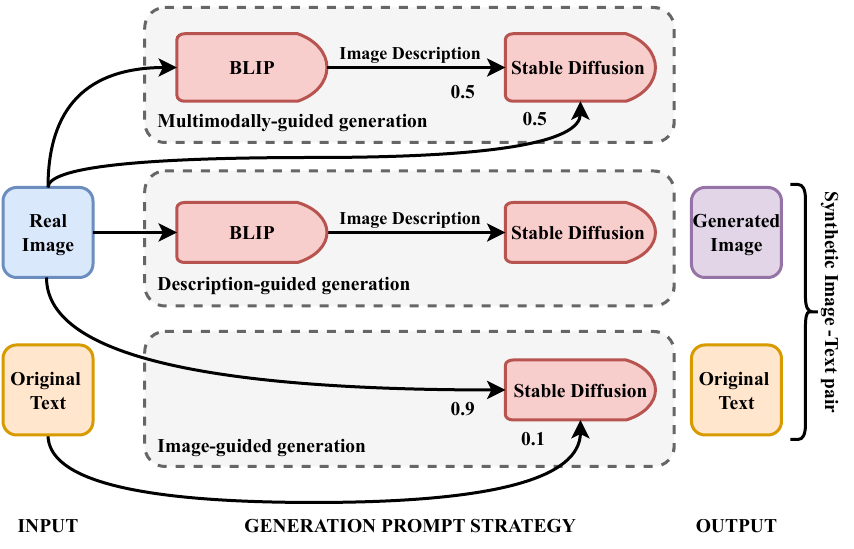}
    \caption{Three prompt strategies used with Stable Diffusion to generate synthetic images based on real images.}
    \label{fig:generationmethods}
\end{figure*}

\subsection{Training Set}
Due to the limited availability of ``in the wild'' AI-generated images, particularly in the \textit{Misinformation} category, we adopted a different strategy for constructing the training set. 
Unlike the evaluation benchmark, which relies on AI-generated images created and shared by actual users, the training set is based on real (non-AI-generated) images related to the three target categories, collected from Reddit and Twitter. 
These were then used to synthesize corresponding AI-generated samples using the Stable Diffusion XL\footnote{\url{https://huggingface.co/stabilityai/stable-diffusion-xl-refiner-1.0}} model.

\subsubsection{Data collection}
For Reddit, we selected subreddits related to \textit{Humor/Satire} or \textit{Art} whose users did not primarily post AI-generated content.
Similarly, for Twitter, we filtered out posts whose Community Notes summaries mentioned generative Artificial Intelligence. 
Nevertheless, any accidentally collected AI-generated images would pose no issue here, as they are still reprocessed into synthetic training samples.
We only included posts whose summaries referenced visual content (e.g., photo, screenshot) to ensure that most retained posts included an image. 
All other filtering criteria were consistent with the evaluation set. 
No suitable art-related community was identified for Twitter in this round of data collection and no misinformation-related subreddit was found in Reddit.
Data collection was conducted in February of 2025. 
As in the case of the test set, the \textit{Misinformation} category was under-represented. 
We applied random under-sampling to the Reddit-derived \textit{Art} and \textit{Humor/Satire} categories, resulting in 29,174 samples per class, as summarized in Table~\ref{tab:trainbal}. 

\subsubsection{Generation Prompt Strategies}
To convert real images into synthetic, we used Stable Diffusion XL to generate AI-synthesized images based on three approaches, as illustrated in Fig. \ref{fig:generationmethods}:
\begin{itemize}
    \item \textbf{Image-guided generation}: The original image was provided as input to the diffusion model (weight: 0.9), along with its original caption (weight: 0.1), simulating image-to-image synthesis with low textual influence.
    \item \textbf{Description-guided generation}: We used BLIP \footnote{https://huggingface.co/Salesforce/blip-image-captioning-large} to caption each original image, then used that description as a text prompt for text-to-image generation.
    \item \textbf{Multimodally-guided generation}: Both the original image and the BLIP-generated caption were used together as inputs, each with equal weight (0.5), enabling balanced multimodal conditioning. 
\end{itemize}
In all cases, the generated image is combined with the text of the initial post to form a training sample. 
This process resulted in three versions of the training dataset, each containing 87,522 generated image–text pairs balanced across the three classes.
The dataset was split into training and validation with a ratio of 80\%-20\% resulting in 70,017 samples for training and 17,505 for validation, balanced across the three classes.

\subsection{Dataset Exploration}
We employed the RINE synthetic image detector \cite{rine}, updated with a DINOv2 backbone, which has shown strong performance on `in the wild' AI-generated images \cite{konstantinidou2025navigatingchallengesaigeneratedimage} and flagged 82\% of \datasetname\ ``in the wild'' samples as synthetic. 
This suggests that the vest majority of images in the test set are AI-generated and detectable as such. 
The remaining images—classified as real—may either originate from more advanced generation models that evade RINE’s detection, or be genuine real images that passed our filtering criteria. 
These findings support our hypothesis that, while general synthetic image detection has seen notable progress, there remains a need for intent-aware detection. 
Simply knowing an image is synthetic provides little insight into whether it was created to mislead, entertain, or express artistic intent.

\begin{figure}[ht!]
    \centering
    \includegraphics[width=0.71\linewidth]{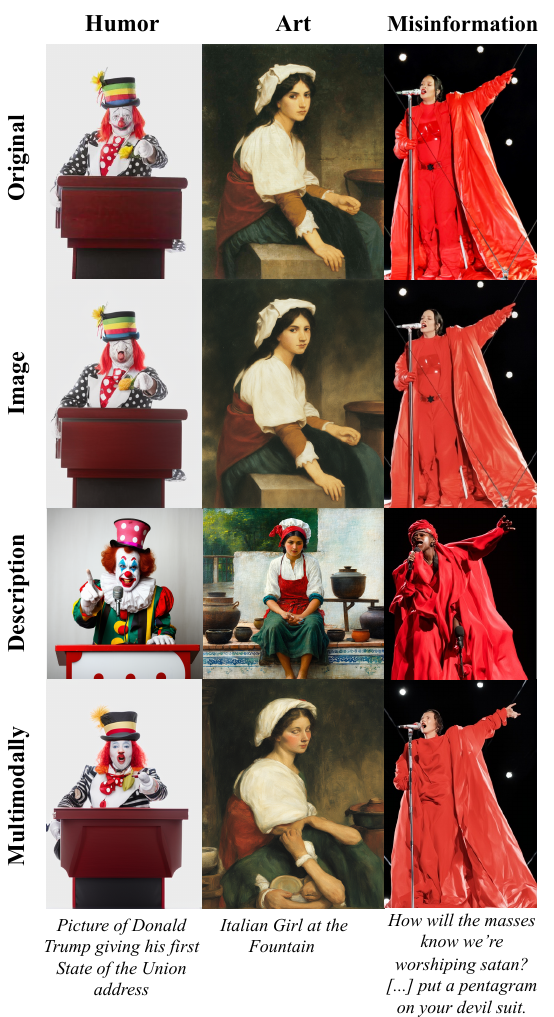}
    \caption{Examples of synthetic images from the three  generation prompt strategies. 
    }
    \label{fig:generationprocedures}
\end{figure}

Figure~\ref{fig:generationprocedures} illustrates examples from the training data generation pipeline. 
The top row displays original, real images collected from Reddit or Twitter. 
Each subsequent row presents AI-generated versions of the originals, using: Image-guided generation (row 2), Description-guided generation (row 3), and Multimodally-guided generation (row 4).
The textual prompts used in the description- and multimodally-guided pipelines were generated using BLIP. The captions for the above examples were: 
(a) ``Clown with red hair and a top hat standing behind a podium with a microphone in his hand and pointing at the camera with a finger up and a smile on his face, with a white background.''
(b) ``Painting of a woman sitting on a ledge with a pot in the background and a pot in the foreground, with a red apron on her head and a white cap on her headband.''
(c) ``Dressed in red singing into a microphone on stage with a red cape on it's head and arms outstretched out to the side of the microphone, with a black background, and lights on a black background.''

We observe that Image-guided generation tends to preserve a high degree of similarity with the original image, primarily altering subtle details such as facial features, softening textures, reducing fine details (as in the art painting), or slightly shifting colors—introducing subtle artifacts characteristic of AI-generated images.
In contrast, Description-guided generation produces the most divergent outputs, while still retaining key conceptual elements (e.g., a clown with red hair and a top hat, a woman sitting on a ledge, a singer wearing red clothes).
These images often exhibit more visible artifacts commonly associated with AI synthesis, such as distorted fingers.
Finally, Multimodally-guided generation strikes a balance between fidelity and alteration—maintaining some visual resemblance to the original while introducing more noticeable changes. 
In this setting, we also observe typical generative artifacts, particularly distorted hands, as seen in the middle image.

\section{Experiments}
\subsection{Experimental Settings}
We employed CLIP ViT-L/14 (quickgelu)\footnote{\url{https://github.com/mlfoundations/open_clip}} \cite{clip} to extract both image $I$ and text embeddings $T$. 
Given input vector $\mathbf{F} \in \mathbb{R}^{d_{\text{in}}}$, we define a classifier consisting of a three-layer Multi-Layer Perceptron (MLP) with GELU activations:
\begin{equation} 
\label{eq:clf}
\mathbf{y} = \mathbf{W}_3 \cdot \mathrm{GELU} \big( \mathbf{W}_2 \cdot \mathrm{GELU}(\mathbf{W}_1 \cdot \mathbf{F}))
\end{equation}
Where $\mathbf{W}_1 \in \mathbb{R}^{512 \times d_{\text{in}}}$, 
$\mathbf{W}_2 \in \mathbb{R}^{128 \times 512}$,  $\mathbf{W}_3 \in \mathbb{R}^{C \times 128}$ and $C = 3$ is the number of output classes. 
Bias terms were included in the network but omitted from the notation for brevity.  
The network is trained using the AdamW optimizer based on the Cross-Entropy Loss. 
Furthermore, we consider multiple modality fusion approaches and detection models to train and evaluate for the task of intent-aware classification:
\subsubsection{Baselines}
We use the MLP classifier and explore unimodal: image-only ($F=I$) and text-only ($F=T$) as well as multimodal experiments, with concatenation ($F=[I;T]$), addition $F=[I+T]$, subtraction $F=[I-T]$, or multiplication $F=[I*T]$
for modality fusion. 
We also consider the combination of the aforementioned fusion approaches, termed CASM for short, first introduced in \cite{papadopoulos2024reddotmultimodalfactcheckingrelevant}, which can be expressed as: $\mathbf{F} = [(I;I+T; I-T; I*T;T)]$.

\subsubsection{Contrastive Learning}
We explore contrastive learning techniques, which have been widely used in multimodal tasks \cite{wu2025generalizablesyntheticimagedetection, clip}, to the image representations using four different loss functions.
The input embedding $I$ was passed through separate projection networks, $\mathbf{z} = \mathbf{W}_{c2} \cdot \mathrm{GELU}(\mathbf{W}_{c1} \cdot \mathbf{x})$, where $\mathbf{W}_{c1} \in \mathbb{R}^{512 \times d_{\text{in}}}$ and $\mathbf{W}_{c2} \in \mathbb{R}^{d_{in} \times 512}$, which are trained with either: Triplet Loss, Quadruplet Loss, InfoNCE Loss, or Supervised Contrastive Loss  \cite{khosla2021supervisedcontrastivelearning}. 
The resulting embeddings were concatenated with the text features and passed to the MLP classifier. 

\subsubsection{Reconstruction Networks}
We investigate the use of Reconstruction Networks, which have been previously applied to tasks such as image super-resolution \cite{yan2024researchimagesuperresolutionreconstruction} and deepfake detection \cite{reconstructiondeepfake}. In our context, inspired by \cite{papadopoulos2025latent}, we aim to leverage reconstruction as a means of restoring latent features of the original image used to generate the synthetic image.
Given the image embeddings $I$ of a synthetic image, we train a neural network to reconstruct the embeddings of the original image $I^o$—that is, the real image from which the synthetic version was derived. The model is trained by minimizing the Mean Squared Error (MSE) between the reconstructed embedding $I^r$ and the target embedding $I^o$.
We explore several architectural variations:
\begin{itemize}
    \item \textbf{Shared Reconstructor (Replace):} A single reconstructor is trained across all categories. The reconstructed embedding $I^r$ replaces $I$, and the classifier receives input $F = [I^r; T]$.
    
    \item \textbf{Shared Reconstructor (Combine):} A single reconstructor is trained across all categories. Both $I$ and $I^r$ are combined with $T$, forming $F = [I; I^r; T]$, which is then passed to the classifier.
    
    \item \textbf{Class-Specific Reconstructors (Replace):} Separate reconstructors $R_h$, $R_a$, and $R_m$ are trained for the humor, art, and misinformation classes, respectively. 
    Only reconstructed embeddings are used, yielding $F = [I^r_h; I^r_a; I^r_m; T]$, as classifier input.
    
    \item \textbf{Class-Specific Reconstructors (Combine):} Same as above, but $I$ is also included in the classifier: $F = [I; I^r_h; I^r_a; I^r_m; T]$.
\end{itemize}

\subsubsection{Attention Mechanisms}
We investigate the use of self-attention mechanisms to capture interactions between image and text modalities. 
In addition, we adapt RED-DOT \cite{papadopoulos2024reddotmultimodalfactcheckingrelevant}, a state-of-the-art Transformer model for multimodal misinformation detection. 
For our purposes, we exclude the Relevant Evidence Detection (RED) module, as it is not applicable in this setting.

\subsubsection{Implementation Details}
To ensure consistency and reliable comparisons, each trainable model was run 10 times using different random seeds, and the results were averaged.
All experiments used a learning rate of $5 \times 10^{-5}$, weight decay of $1 \times 10^{-4}$, and a batch size of 32.
Each model was trained for up to 30 epochs.

\subsubsection{Large Vision-Language Models}
We explore the use of LVLMs for zero-shot intent-aware classification of image-text pairs.
Specifically, we use Llama-3.2-11B\footnote{https://huggingface.co/meta-llama/Llama-3.2-11B-Vision-Instruct}.
Given the image-text pair as input, we use the ``Direct Classification'' prompt:
\begin{tcolorbox}[colback=white, colframe=black, title=``Direct Classification'' prompt]
\textit{``You are given a title and an AI generated image. Your task is to classify the pair into one of the following categories: `misinformation', `satire', or `art'. 
Respond with only the category name, no punctuation, no explanation, and no additional text.''}
\end{tcolorbox}
Due to high confusion between satire and misinformation, we also added the \textit{``Between `misinformation' and `satire' choose `misinformation''} sentence; denoted as ``Direct Classification (Nudged)''

Furthermore, we explore a ``Two-Stage Prompting'' approach, in which image analysis is performed before classification.
In the first stage, the image is provided as input along with the following prompt:
\begin{tcolorbox}[colback=white, colframe=black, title=First-stage Prompt (Description)]
\textit{``You are given an AI-generated image. Please describe in detail what the image contains. Describe the emotional tone conveyed. Respond with a detailed analysis, including what stands out and why.''}
\end{tcolorbox}
In Stage 2, the image, original caption, and the description from Stage 1 are passed together with the following prompt:
\begin{tcolorbox}[colback=white, colframe=black, title= Second-stage Prompt (Detection)]
\textit{``You are given a title, an AI-generated image, and a detailed analysis of the image. Your task is to classify the image-text pair into one of the following categories: `misinformation', `satire', or `art'. Between `misinformation' and `satire' choose `misinformation'. 
Respond with only the category name, no punctuation, no explanation, and no additional text.''}
\end{tcolorbox}

\begin{table*}
  \caption{
  Comparison of detection models trained on datasets from three generation prompt strategies. 
  We report the average accuracy and standard deviation over 10 random seeds. 
  \textbf{Bold} indicates the best overall performance; \underline{underlining} highlights the best result per generation prompt strategy.
  }
  \label{tab:experiment_results}
  \centering
  \begin{tabular}{lccc}
    \toprule
    \textbf{Method / Variation} & \multicolumn{3}{c}{\textbf{Accuracy \% (Std. Dev. \%)}} \\
    \cmidrule(lr){2-4}
     & \textbf{Image-guided} & \textbf{Description-guided} & \textbf{Multimodally-guided} \\
    \midrule
    MLP (Image only) & 62.94 (1.33) & 61.62 (1.02) & 63.80 (0.68) \\
    MLP (Text only) & 65.43 (0.45) & 65.43 (0.45) & 65.43 (0.45) \\
    \midrule
    MLP (Concatenation) & \textbf{\underline{\text{71.60}}} (0.54) & 70.05 (0.53) & 70.32 (0.39) \\
    MLP (Addition) & 69.61 (0.62) & 68.37 (0.70) & 68.10 (0.97) \\
    MLP (Subtraction) & 65.47 (0.46) & 62.29 (0.92) & 65.02 (0.69) \\
    MLP (Multiplication) & 59.69 (1.04) & 57.13 (0.72) & 59.26 (1.02) \\
    MLP (CASM) & 71.37 (0.89) & 70.05 (0.71) & 70.28 (0.53) \\
    \midrule
    Contrastive Triplet Loss & 69.51 (0.96) & 69.94 (0.59) & 70.61 (0.86) \\
    Contrastive Quadruplet Loss & 68.67 (0.68) & 69.43 (0.49) & 69.87 (0.60) \\
    Contrastive InfoNCE & 70.88 (0.53) & 68.79 (0.88) & 70.39 (0.45) \\
    Supervised Contrastive & 70.03 (0.65) & 70.13 (0.49) & \underline{\text{71.50}} (0.57) \\
    \midrule
    Shared Reconstructor (Replace) & 70.54 (0.80) & \underline{\text{70.77}} (0.58) & 71.14 (0.64) \\
    Shared Reconstructor (Combine) & 70.75 (0.79) & 69.94 (0.81) & 69.89 (1.07) \\
    Class-specific Reconstructors (Replace) & 69.64 (1.96) & 69.22 (1.41) & 68.92 (1.42) \\
    Class-specific Reconstructors (Combine) & 71.16 (0.76) & 69.79 (1.17) & 70.23 (1.44) \\
    \midrule
    Self-Attention & 70.05 (0.31) & 69.37 (0.23) & 68.90 (0.32) \\
    RED-DOT & 70.05 (1.04) & 69.25 (0.93) & 70.16 (0.90) \\
    \midrule
    \rowcolor{gray!20} Average Accuracy & 68.67 (0.81) & 67.74 (0.74) & 68.46 (0.73) \\
    \bottomrule
  \end{tabular}
\end{table*}

\subsection{Experimental Results}

\subsubsection{Comparative study across generation prompt strategies}
Table~\ref{tab:experiment_results} presents the performance of all methods trained on the synthetic \textit{\datasetname} training set and evaluated on the ``in the wild''  benchmark. 
We observe that the highest overall performance is achieved by the MLP classifier with concatenated image and text features ($F = [I;T]$), reaching 71.6\% accuracy with a standard deviation of 0.54 when trained on image-guided data. 
Among contrastive methods, pre-training with supervised Contrastive Loss reached the second best performance (71.5\%) when trained on multimodally-guided data. 
Meanwhile, among reconstruction-based approaches, the Class-Specific Reconstructors (Combine) performed well when trained on image-guided data, with an average accuracy of 71.16\%, ranking fourth across all methods.
We also observe that attention-based models underperformed compared to simpler architectures. 
Self-attention peaked at 70.05\% on image-guided data, and RED-DOT reached 70.16\% with multimodally-guided data; trailing behind the ``MLP (CASM)'' baseline without any attention mechanism.

We observe that the three prompting strategies--image-guided, description-guided, and multimodally-guided--favor different model architectures, suggesting that each introduces distinct patterns or artifacts that certain models are better equipped to exploit. 
As shown in the last row of Table~\ref{tab:experiment_results}, the description-guided approach consistently underperforms, likely due to the absence of direct visual input during generation, resulting in missing visual cues essential for classification.
In contrast, the image- and multimodally-guided methods retain these visual features, supporting stronger generalization across models.

\subsubsection{Performance of Large Language-Vision Models}
As shown in Table~\ref{tab:lvlmexpt}, LVLMs underperform compared to task-specific models. 
This is expected, as LVLMs operate in a zero-shot setting without fine-tuning for the classification task.
The baseline Direct Classification prompt yields limited accuracy (50.09\%). 
However, the `Nudged' prompt--designed to bias the model toward selecting misinformation when uncertain between satire and misinformation--leads to a substantial improvement, reaching 62.28\%. 
Moreover, the two-stage prompting strategy, which first elicits a description of the image before classification, further increases accuracy to 66.65\%.
While LVLMs still fall short of fully trained models, these findings indicate promising directions for improvement, such as few-shot prompting, more advanced models, and the integration of chain-of-thought reasoning. 
This suggests that with appropriate prompting strategies, LVLMs may become viable components in future intent-aware and multimodal misinformation detection systems.

\subsubsection{Model Generalization}
Table~\ref{tab:perclasssource} reports per-class accuracy for the best-performing method, MLP (concatenation) trained on image-guided data, across the validation set and ``in the wild'' data from Twitter and Reddit. 
First, we observe that accuracy on the validation set is notably high, averaging 96.63\% across the three categories. 
Nevertheless, this this performance does not directly translate to ``in the wild'' synthetic images. 
This gap highlights the distributional shift between our synthetic training data and ``in-the-wild'' AI-generated content, which is more diverse, produced by a range of generative models, and curated by users for quality.

Performance on the real-world benchmark is especially low for satire/humor content from Twitter (13.56\%), which corresponds with the under-representation of this class–source pair in the dataset, as shown in Tables~\ref{tab:testtotalbal} and~\ref{tab:trainbal}. 
The resulting data imbalance likely hampers the model’s ability to learn meaningful patterns.
In contrast, the \textit{Art} and \textit{Misinformation} classes achieve higher accuracy on Twitter (72.71\% and 85.59\%, respectively), likely due to the longer average text length of Twitter posts compared to Reddit, offering richer textual context and aiding classification. 

\begin{table}
  \caption{Zero-shot performance of LVLM-based classification.}
  \label{tab:lvlmexpt}
  \centering
  \begin{tabular}{lc}
    \toprule
    \textbf{Prompt} & \textbf{Accuracy} 
    \\
    \midrule
    Direct Classification & 50.09 
    \\
    Direct Classification (Nudged) & 62.28 
    \\
    Two-Stage (Nudged) & 66.65 
    \\
    \bottomrule
  \end{tabular}
\end{table}

\begin{table}
  \caption{Per-class accuracy of MLP (Concatenation) on validation and `in the wild' Twitter and Reddit samples.}
  \label{tab:perclasssource}
  \centering
  \begin{tabular}{lccc}
    \toprule
    \textbf{Category} &  \textbf{Validation set} & \textbf{Twitter} & \textbf{Reddit} \\
    \midrule
    Humor/Satire  & 94.61 & 13.56 & 69.59  \\    
    Art & 99.53 & 72.71 & 61.36  \\
    Misinformation & 95.75 & 85.59 & - \\
    \midrule
    \rowcolor{gray!20} Balanced Accuracy & 96.63 & 75.49 & 66.57 \\
    \bottomrule
  \end{tabular}
\end{table}

\subsubsection{Unimodal Analysis}
As shown in Table~\ref{tab:experiment_results}, text-only models consistently outperform image-only models across all three generation procedures. 
This suggests that text provides more informative signals for classification, likely because synthetic images--regardless of their intended purpose--are often visually similar. 
In contrast, captions offer stronger cues about the creator's intent.
Unsurprisingly, the text-only results remain identical across generation types, since all methods retain the original, unmodified captions.
Nevertheless, models trained on both modalities consistently outperform unimodal counterparts, highlighting the importance of cross-modal interactions for accurate classification; validating our hypothesis that intent-aware classification of synthetic images should be addressed as a multimodal task. 

\section{Conclusion}
In this work, we introduced \textit{\datasetname}, a novel dataset for intent-aware classification of synthetic multimodal content. Unlike previous efforts that focus on detecting synthetic or decontextualized images, \textit{\datasetname} targets a more nuanced challenge: understanding the intent behind AI-generated images by categorizing them as humor/satire, art, or misinformation.
\textit{\datasetname} includes a real-world ``in-the-wild'' evaluation benchmark sourced from Twitter and Reddit, alongside three synthetically generated training sets produced using Stable Diffusion through image-guided, description-guided, and multimodally-guided generation strategies.

We conducted a comparative study on a wide range of models, including various modality fusion techniques, contrastive learning, reconstruction networks, attention mechanisms and LVLMs. 
These methods achieved limited model generalization, highlighting the difficulty of inferring intent—a high-level, abstract concept—from visual and textual cues alone.
Our findings indicate that current architectures are ill-suited for intent-aware classification, highlighting the need for models that can reason about intent, contextual framing, and social subtext--alongside more diverse and representative training data.
We hope that \textit{\datasetname} will serve as a foundation for advancing research in this direction.

\section*{Acknowledgments}
This work is partially funded by the Horizon Europe projects ``DisAI - Improving scientific excellence and creativity in combating disinformation with artificial intelligence and language technologies'' under grant agreement no. 101079164, 
``vera.ai - VERification Assisted by Artificial Intelligence'' under grant agreement no. 101070093, 
and ``AI-CODE - AI services for COntinuous trust in emerging Digital Environments'' under grant agreement no. 101135437.
The authors also acknowledge the support and computational resources provided by the IT Center of the Aristotle University of Thessaloniki (AUTh) throughout the progress of this research work.

\balance
\bibliographystyle{unsrt}  
\bibliography{main}

\end{document}